\pdfoutput=1

\documentclass[11pt]{article}

\usepackage{natbib}
\setcitestyle{numbers,square}
\usepackage{CUSTOM}

\usepackage{times}
\usepackage{latexsym}

\usepackage[T1]{fontenc}

\usepackage[utf8]{inputenc}

\usepackage{microtype}

\usepackage{inconsolata}

\usepackage{algorithm}
\usepackage{algorithmic}
\usepackage{amsmath}
\usepackage{graphicx}
\usepackage{multirow}
\usepackage{subcaption}

\graphicspath{ {./} }

\title{Recognition of Mental Adjectives in An Efficient and Automatic Style}

\author{Fei Yang \\
         MathAI Lab \\
         yftadyz@163.com}

\begin{document}
\maketitle
\begin{abstract}
In recent years, commonsense reasoning has received more and more attention from academic community. We propose a new lexical inference task, \emph{Mental} and \emph{Physical} Classification (MPC), to handle commonsense reasoning in a reasoning graph. \emph{Mental} words relate to mental activities, which fall into six categories: Emotion, Need, Perceiving, Reasoning, Planning and Personality. \emph{Physical} words describe physical attributes of an object, like color, hardness, speed and malleability. A BERT model is fine-tuned for this task and active learning algorithm is adopted in the training framework to reduce the required annotation resources. The model using ENTROPY strategy achieves satisfactory accuracy and requires only about 300 labeled words. We also compare our result with SentiWordNet to check the difference between MPC and subjectivity classification task in sentiment analysis.
\end{abstract}

\section{Introduction}
\begin{figure}[h!]
  \includegraphics[scale=0.3]{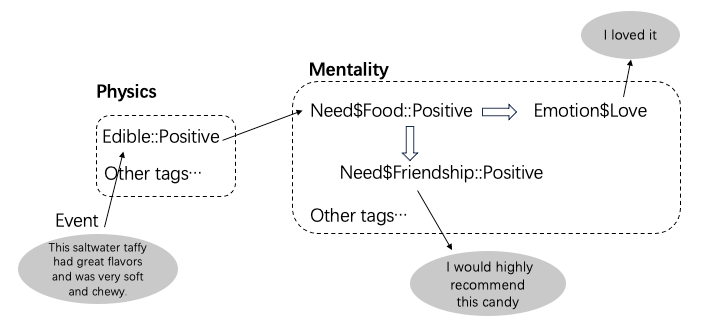}
  \caption{A reasoning graph between a physical event and mental reactions. \emph{Edible::Positive} means positive effect over a physical attribute \emph{Edible}. In mentality part, \emph{Need\$Food::Positive}, \emph{Need\$Friendship::Positive} mean positive effect over \emph{Food} and \emph{Friendship} respectively, which belong to \emph{Need} category \citep{maslow1943theory}. \emph{Need\$Food::Positive} invokes \emph{Love} belonging to \emph{Emotion} category \citep{plutchik1980general}. Other tags not invoked are omitted. }
  \label{fig:reasoning_path}
\end{figure}

In the field of artificial intelligence, commonsense reasoning refers to the capacity that a machine understands the nature of scenes commonly encountered by humans every day, and makes reasonable and appropriate reactions, mimicking human cognitive abilities. Through commonsense reasoning, humans are capable of intricate reasoning relating to fundamental domains including time, space, naive physics, and naive psychology \citep{davis2015commonsense}. Therefore, a good starting point is understanding how time, space, naive physics affect human's mind, exploring possible causal relationships. For example, let's consider a review "This saltwater taffy had great flavors and was very soft and chewy. I loved it and I would highly recommend this candy!". The concepts "great flavors", "soft", "chewy" describe physical attributes of the saltwater taffy and the concepts "love", "recommend" describe mental activities of the reviewer. Here concept refers to word or phrase in natural language. If a seven years old child reads this review, the child would understand that the mental activities are caused by the taffy's physical attributes. Figure~\ref{fig:reasoning_path} shows a possible reasoning graph existed in the child's mind. The words "great flavors", "soft", "chewy" indicate that this taffy is edible with a positive effect. This effect greatly satisfies the reviewer's need of food and then this strong satisfaction invokes the reviewer's emotion of love with an reaction "I love it". That strong satisfaction also invokes the reviewer's need of friendship positively, with an reaction that the reviewer would like to share this taffy with friends. 

To let a machine figure out a similar reasoning graph, the first step is recognizing which concept is \emph{Physical} and which one is \emph{Mental}. Then all concepts are mapped to numerous more granular tags, like \emph{Edible::Positive} or \emph{Need\$Friendship::Positive} shown in Figure~\ref{fig:reasoning_path}. Last, all tags are linked together to form a powerful reasoning graph. The first step cannot be skipped because the coarsest reasoning path, \emph{Physical} -> \emph{Mental}, provides causal concept pairs, facilitating design of more fine-grained tags. Under this research plan, we propose a task of \emph{Mental} and \emph{Physical} Classification (MPC) at lexical level in this work. Each adjective extracted from Amazon Fine Food Reviews dataset \citep{mcauley2013amateurs} is inferred with a binary tag, \emph{Mental} or \emph{Physical}, by a fine-tuned BERT model. A \emph{Mental} adjective describes mental activities, like emotion, need, reasoning, while a \emph{Physical} one shows physical attributes of an object, like color, hardness, speed and malleability. Although our inference methods have been for adjectives, they can be directly applied to other word classes.  The inferred tags of MPC only reveals that an adjective is more likely to express a mental view or a physical view, as a word might have different senses. 

Besides MPC, dozens of binary or multi-value tags, like \emph{Emotion} or \emph{Need} category will be developed in the follow-up research work. Moreover, in order to improve the reasoning performance, these tags might need to be updated or new tags join the reasoning graph. This continuous and rapid iterative process makes it impossible to annotate all the words at once. In fact, what this project really needs is the ability to tag all the words automatically relying on zero or very low annotation resources. Therefore, we consider active learning methods to train a BERT \citep{devlin2018bert}  model for MPC in this work. ENTROPY \citep{lewis1995sequential}, CORESET \citep{sener2017active}, CAL \citep{margatina2021active} and Random strategies are implemented and evaluated. The experiment results indicate that ENTROPY outperforms others and achieves \emph{Mental} F1 0.72 and \emph{Physical} F1 0.87 on testset, with only around 300 words are annotated for training. 

The definition of MPC task bears some similarity to subjectivity classification which is one task of sentiment analysis \citep{liu2010sentiment}, and classify whether a piece of text is objective or subjective. To investigate the difference between these tasks, our result by ENTROPY is compared with SentiWordNet \citep{sebastiani2006sentiwordnet, baccianella2010sentiwordnet}. We find that  41.5\% of the \emph{Mental} adjectives bear objective meanings, which indicates the notion of MPC is quite different from subjectivity classification. Adjective examples are listed out to illustrate this difference in Table~\ref{tab:ob_word_mental_func2}. 

The main contributions of this paper include the following three points: (1) a new task MPC is proposed to handle commonsense reasoning, (2) active learning is introduced to solve MPC efficiently, relying on only a small size of annotated words, (3) a dataset with the inferred MPC tags is released publicly for future research.

\section{Related Work}
\textbf{Commonsense Reasoning. } Reasoning between mentality and physics has been studied by the research community in recent years. The mental reason of affective events is explained based on seven common human needs \citep{Ding2018WhyIA}. Event2Mind studies two kind of mental state, intent and emotion, which are inferred by deep learning models given physical events described by short text-free phrases \citep{Rashkin2018Event2MindCI}. ATOMIC considers two more kind of mental state, planning and personality, under the same task setting of Event2Mind \citep{Sap2019ATOMICAA}.  Reasoning between physical events are studied by \citep{Zellers2019HellaSwagCA} and \citep{Talmor2019CommonsenseQAAQ}. Previous works provide no clear explanation about "how" and "why" in commonsense reasoning, which is the core question that our research works try to address.

\textbf{Sentiment Analysis. } 
Subjectivity classification and sentiment classification are two sub-topics of sentiment analysis \citep{liu2010sentiment}.  Subjectivity classification is to determine whether a content is objective or subjective. On the other hand, sentiment classification is utilized for subjective content to identify the sentiment polarity, that is, whether the author expresses a positive or negative opinion.  One approach to sentiment analysis is using lexicons where each word is assigned with scores showing it is neutral, positive or negative \citep{taboada2011lexicon, hatzivassiloglou1997predicting, hu2004mining, taboada2006methods}. These scores are known as prior polarity, that is, irrespective of the context, whether the word convey a positive or negative or neutral connotation \citep{wilson-etal-2005-recognizing}. One popular lexical resource is SentiWordNet \citep{sebastiani2006sentiwordnet, baccianella2010sentiwordnet} which associates polarity scores to each synset of WordNet \citep{miller1995wordnet}. Early researches in this domain focus on adjectives, as adjectives express the majority of subjective meaning in a piece of writing \citep{hu2004mining, taboada2006methods}. Under the same consideration, we also focus on adjectives for MPC first in this work. 

\textbf{Active Learning. } When machine learning or deep learning algorithms are considered to solve NLP tasks, one of most common challenges is lack of labeled data and limited annotation resources due to project budget. To efficiently make use of annotation resources, only the most valuable samples are hoped to be selected out for human labeling. Active learning provides a set of algorithms to fulfill this goal \citep{settles2009active}. ENTROPY is an uncertainty-based method, choosing the sample with the highest predicted entropy \citep{lewis1995sequential}.  However, the problem with this approach is that there is a risk of picking outliers or similar samples \citep{settles2009active}. To increase diversity of the selected samples, CORESET \citep{sener2017active} chooses the furthest sample in the embedding space from the samples already selected in previous iterations.  CAL \citep{margatina2021active} finds the most contrastive sample to its nearest neighbors by calculating KL divergence, leveraging both uncertainty and diversity. 

\textbf{BERT. } In recent years BERT \citep{devlin2018bert} has become one of the most famous pre-training language models and has shown effectiveness in many natural language processing tasks. These include sentiment analysis \citep{socher-etal-2013-recursive}, semantical similarity \citep{dolan2005automatically}, question answering \citep{rajpurkar-etal-2016-squad} and entailment inference \citep{williams2017broad}. BERT is pre-trained on the BooksCorpus (800M words) \citep{zhu2015aligning} and English Wikipedia (2,500M words). By pre-training on such large text data, BERT grasps rich semantic information. The most common usage of BERT is fine-tuning it over downstream tasks, trained with data from downstream tasks to update all its pre-trained parameters. By this way, both the rich semantic information from pre-training and the features from downstream tasks are taken advantage of to achieve an excellent performance.

\section{Data and Annotation}
\label{sec:da}
\textbf{Task Definition. } In this work, we define a binary classification task, inferring a word is \emph{Mental} or \emph{Physical}. The notion of \emph{Mental} relates to mental activities, which fall into six categories: Emotion, Need, Perceiving, Reasoning, Planning and Personality. Personality are regarded as the external manifestation of persistent mental activities. Detailed definition of each category and word examples are shown in Table~\ref{tab:ob_word_mental_func2}. Other words are defined as \emph{Physical} describing physical attributes of an object, like color, hardness, speed and malleability. \emph{Mental} words usually have abstract meanings, but \emph{Physical} words have more concrete meanings that can be observed in the world. This difference can be used as a simple reference to determine which class a word belongs to. The inferred class only reveals that a word is more likely to express a mental view or a physical view, as a word might have different senses. The main reason we choose lexical level rather than sense level for MPC, is to facilitate subsequent research and reduce development complexity.

\textbf{Data Process. } Amazon Fine Food Reviews dataset \citep{mcauley2013amateurs} \footnote{This dataset is distributed under CC0: Public Domain License. Download url: \url{https://www.kaggle.com/datasets/snap/amazon-fine-food-reviews}} is used as corpus for MPC task, as this dataset contains reasoning between physics (food description) and mentality (people's opinion) in our daily life. It has more than 0.5 million reviews of Amazon fine foods from Oct 1999 to Oct 2012. We use only text column and remove all other columns like ProductId, UserId, ProfileName for anonymization considerations. Data process contains three steps. An processing example is given in Table~\ref{tab:review_process}. First, each piece of review is splitted into words and each word is classified into part of speech (POS-tagging) \footnote{Penn Treebank POS tags are used. See details in \url{https://www.ling.upenn.edu/courses/Fall_2003/ling001/penn_treebank_pos.html}}. Then (adjective, noun) pairs are recognized and extracted out, where the noun appeared immediately after the adjective in review sentences. The goal of this step is to make sure the extracted adjectives are used in daily life to describe objects or states, like "roasted beans", "angry complaint", and counteract possible errors from POS-tagging. Finally, adjectives from these pairs are validated by checking if they have definition text from WordNet. After deduplication, 7292 adjectives are obtained. We use version 3.6.7 of NLTK package for POS-tagging and WordNet calling. 

\begin{table}
\centering
\begin{tabular}{l}
\hline
\begin{tabular}[c]{@{}l@{}} \textbf{Review}: I have found them all to be\\ of good quality. \end{tabular} \\
\hline
\begin{tabular}[c]{@{}l@{}}  \textbf{Step 1}: Pos-tagging. \\\textbf{Result}: ("I", "PRP"), \\ ("have", "VBP"), ("found", "VBN"), \\ ("them", "PRP"), ("all", "DT"), ("to", "TO"),\\ ("be", "VB"), ("of", "IN"), ("good", "JJ"), \\("quality", "NN"), (".", ".")  \end{tabular} \\
\hline
\begin{tabular}[c]{@{}l@{}}  \textbf{Step 2}: Detect (adjective, noun) pairs. \\\textbf{Result}: ("good", "quality")  \end{tabular} \\
\hline
\begin{tabular}[c]{@{}l@{}}  \textbf{Step 3}: Validate adjectives. \\ \textbf{Result}: "good"  \end{tabular} \\
\hline
\end{tabular}
\caption{\label{tab:review_process}
Review process pipeline. Input is "I have found them all to be of good quality." and after processing the word "good" is outputted.
}
\end{table}

\textbf{Annotation. }  Each adjective is annotated by two annotators checking word definition from WordNet, and disagreements are adjudicated by another expert. All participants are experienced volunteers and they are notified how their annotations are used in this work. Examples of words and their definitions are presented in Appendix~\ref{sec:appendix_infersub_example}. For words with different senses, annotation results are mainly based on the frequency of daily use. For instance, although the word "cold" has a \emph{Mental} sense, "feeling or showing no enthusiasm", it's labeled as \emph{Physical} since it is used more frequently with the gloss "having a low or inadequate temperature or feeling a sensation of coldness or having been made cold by e.g. ice or refrigeration". 

A testset consisting of 100 words is annotated for measuring model performance. It contains 26\% \emph{Mental} words and 74\% \emph{Physical} words. Among the \emph{Mental} words, 12\% of them have annotation disagreements while this number drops to 5\% for \emph{Physical} words. This difference indicates that \emph{Mental} words are more likely to be misclassified. Total disagreement over this dataset between two annotators is 7\% . Statistics of the testset is summarized in Table~\ref{tab:testset_stat}. For each active learning strategy, a dataset for training and validation is annotated, which has no overlap with the testset.

\begin{table}
\centering
\begin{tabular}{cccc}
\hline
\textbf{Class} & \textbf{Total} & \textbf{Disagreement} & \textbf{Rate} \\
\hline
\emph{Mental} & 26 & 3 & 12\% \\
\emph{Physical} & 74 & 4 & 5\% \\
\hline
\end{tabular}
\caption{\label{tab:testset_stat}
Total word numbers, disagreement numbers and rate of disagreement of the two classes in the testset. Difference of disagreement rates indicates that \emph{Mental} words are more likely to be misclassified.
}
\end{table}

\section{Methods}
We use active learning framework to train a binary classifier for MPC task, which is shown in Algorithm \ref{alg:ac}. An unlabeled word pool $\mathcal{U}$ is set up consisting of the extracted adjectives. The random strategy is used to select a word for annotation in the first iteration, while in other iterations different active learning strategies are used. We aim to annotate $K_1$ positives and $K_2$ negatives in each iteration, which are put into a labeled word pool $\mathcal{D}_{labeled}$. A threshold $M$ is set to control the total number of annotation of each iteration, in case that the active learning strategy fails to find another positive or negative sample. At the end of each iteration, a BERT model is fine-tuned over $\mathcal{D}_{labeled}$. When iterations end, the BERT model with best performance over testset is employed in pipeline for inference. 

\begin{algorithm}
\caption{Active Learning Framework}
\label{alg:ac}
\begin{algorithmic}[1]
\REQUIRE Unlabeled word pool $\mathcal{U}$, number of positive samples $K_1$ and negative samples $K_2$ and maximum annotated samples $M$ per iteration, number of iterations $T$
\STATE $\mathcal{D}_{labeled} = \{\}$
\STATE $t = 0$
\WHILE { $t < T$ }
	\STATE $\mathcal{D}_{pos}, \mathcal{D}_{neg} = \{\}, \{\}$
	\STATE $m = 0$
	\WHILE {True}
		\IF {$t = 1$}  
			\STATE $w_{new} \gets$ Randomly select a word from $\mathcal{U}$  
		\ELSE
			\STATE $w_{new} \gets$ Select a word from $\mathcal{U}$ by a specific strategy
		\ENDIF
		\STATE Annotate $w_{new}$ with a class label $C$
		\STATE $\mathcal{U} = \mathcal{U} \setminus \{ w_{new} \}$
		\STATE $m=m+1$
		
		\IF {$C$ is positive and $|\mathcal{D}_{pos}| < K_1$}
			\STATE $\mathcal{D}_{pos} = \mathcal{D}_{pos} \cup \{(w_{new}, C)\}$
		\ENDIF
		\IF {$C$ is negative and $|\mathcal{D}_{neg}| < K_2$}
			\STATE $\mathcal{D}_{neg} = \mathcal{D}_{neg} \cup \{(w_{new}, C)\}$
		\ENDIF
		\IF {($|\mathcal{D}_{pos}| = K_1$ and $|\mathcal{D}_{neg}| = K_2$) or $m=M$}
			\STATE break
		\ENDIF
	\ENDWHILE
	
	\STATE $\mathcal{D}_{labeled} = \mathcal{D}_{labeled} \cup \mathcal{D}_{pos} \cup\ \mathcal{D}_{neg}$
	\STATE Fine-tune a BERT over $\mathcal{D}_{labeled}$ 
	\STATE $t = t + 1$
	
\ENDWHILE
\end{algorithmic}
\end{algorithm}

\begin{figure}[h!]
  \includegraphics[scale=0.18]{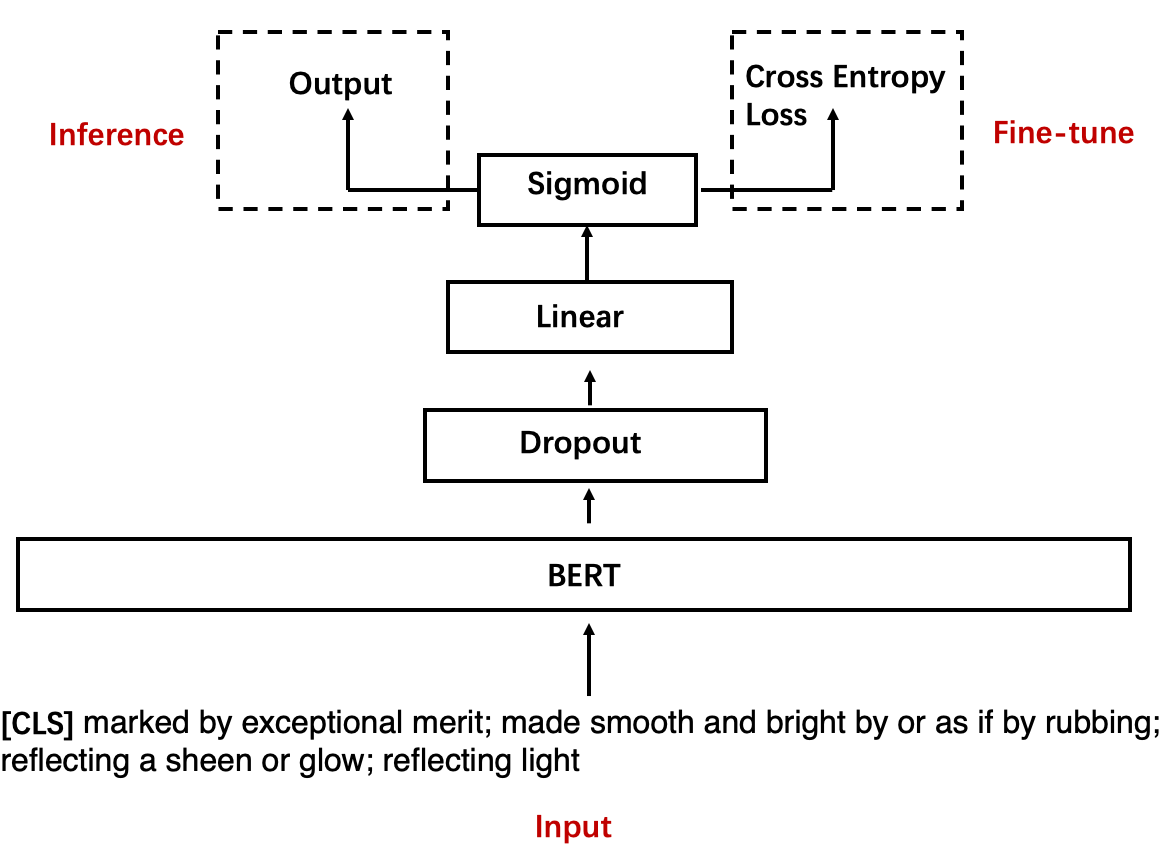}
  \caption{Definitions of words are concatenated with \emph{[CLS]} at head as input of BERT. A dropout and a linear layer are connected to BERT sequentially. At last, a sigmoid node outputs the probability of being positive. The output probability is consumed for inference, or as input of cross entropy loss for fine-tuning. We use a word "shining" with its definition text as an input example. }
  \label{fig:bert-workflow}
\end{figure}

BERT fine-tuning and inference procedure is shown in Figure~\ref{fig:bert-workflow}. As WordNet maps words into sets of cognitive synonyms, each expressing a distinct concept, therefore more than one piece of definition text are provided by WordNet for a given word. For example, "shining" belongs to three clusters as an adjective with three different definitions: (1) marked by exceptional merit, (2) made smooth and bright by or as if by rubbing; reflecting a sheen or glow, and (3) reflecting light. All of them are aggregated as one piece of text, serving as input of BERT with a special token \emph{[CLS]} at head. We use the final hidden state of \emph{[CLS]} as BERT output, which is then connected with a dropout layer \citep{srivastava2014dropout} and a linear layer. Sigmoid node is added after the linear layer to transform logits into the probability of positive class. For fine-tuning, a standard cross entropy loss is computed to update all parameters of the BERT model and the subsequent linear layer.

\section{Experiments and Results}

We compare four active learning strategies, considering their classification performance and annotation resource consumption, to find which strategy is most suitable for MPC task given a limited project budget. \emph{Mental} class serves as positive and \emph{Physical} serves as negative in training. F1 scores of \emph{Mental} and \emph{Physical} classes over testset are computed respectively. The average of the number of labeled samples in each iteration is recorded.

\textbf{ENTROPY. }   Samples with the highest predicted entropy are selected \citep{lewis1995sequential}. For binary classification, the closer the prediction probability of a sample is to 0.5, the higher its entropy is. Therefore, at the starting of each iteration, the BERT model from last iteration outputs the probability of all the words in $\mathcal{U}$ and the word whose probability is closest to 0.5 is selected.

\textbf{CORESET. }   Samples that are furthest away from the samples selected in previous iterations are chosen to enlarge the semantic diversity \citep{sener2017active}. FastText \citep{bojanowski2016enriching} is used to represent a word by embedding vector, as fastText works well in word-level semantic textual similarity (STS) tasks \citep{Zhelezniak2019CorrelationCA}. In each iteration, a word is selected as follows:

\begin{equation}
\label{eq:coreset}
	w_{new} = \mathop{\arg\max_{w \in \mathcal{U}}}{\mathop{\min_{v \in \mathcal{D}_{labeled}}}{L_2( \phi(w), \phi(v))}}, 
\end{equation}

where $L_2(\cdot, \cdot)$ computes $L_2$ distance between two vectors and $\phi(\cdot)$ returns the embedding vector of a word.
 
 The most contrastive sample to its nearest neighbors by calculating KL divergence is chosen.

\textbf{CAL. }   The most contrastive sample to its nearest neighbors by calculating KL divergence is chosen \citep{margatina2021active}. Given a word $w$ in unlabeled word pool $\mathcal{U}$, nearest 10 words are selected as neighbors in labeled word pool $\mathcal{D}_{labeled}$ by $L_2$ distance. The average KL divergence between $w$ and its neighbors is computed as a measure of contrastive degree. The word with the largest value of this measure is selected. 

\textbf{Random. }   Select a word $w$ in unlabeled word pool $\mathcal{U}$ randomly.

\begin{table*}[ht]
    \begin{subtable}{.5\linewidth}
      \centering
        \caption{\emph{Mental} F1}
        \resizebox{!}{0.8cm}{
        \begin{tabular}{ccccc}
            \hline
            \textbf{Iteration} & \textbf{ENTROPY} & \textbf{CORESET} & \textbf{CAL} & \textbf{Random} \\
            \hline
             3 & 0.61 & 0.69 & 0.62 & 0.61 \\
             4 & \textbf{0.72} & \textbf{0.71} & 0.64 & 0.64 \\
             5 & \textbf{0.70} & 0.69 & 0.68 & 0.64 \\
            \hline
        \end{tabular}
        }
    \end{subtable}
    \begin{subtable}{.5\linewidth}
      \centering
        \caption{\emph{Physical} F1}
        \resizebox{!}{0.8cm}{
        \begin{tabular}{ccccc}
            \hline
            \textbf{Iteration} & \textbf{ENTROPY} & \textbf{CORESET} & \textbf{CAL} & \textbf{Random} \\
            \hline
            3 & 0.82 & 0.66 & 0.79 & 0.71 \\
             4 & \textbf{0.87} & \textbf{0.83} & 0.80 & 0.66 \\
             5 & \textbf{0.85} & 0.81 & 0.76 & 0.69 \\
	    \hline	
        \end{tabular}
        }
    \end{subtable} 
    \caption{Averaged F1 scores after 3,4,5 iterations. ENTROPY outperforms the other three, achieving the highest \emph{Mental} F1 0.72 and \emph{Physical} F1 0.87 at iteration 4.} 
    \label{tab:f1}
\end{table*}

All strategies share the same experimental settings as following: total iterations $T$ = 5, number of positive samples $K_1$ = 20, number of negative samples $K_2$ = 20, maximum annotation number $M$ = 120. In each iteration, BERT fine-tuning takes totally 20 epochs with learning rate 2e-5 and batch size 32. Learning rate drops to 1/10 of the original level after 10 epochs. We split $\mathcal{D}_{labeled}$ by 80\% - 20\% as trainset and devset. If BERT outputs a value greater than 0.5, the word is considered to belong to \emph{Mental}, otherwise \emph{Physical}. Winner model $\mathcal{M}_t$ is the one with maximum accuracy over devset. We hypertune BERT with different values of learning rate \{1e-5, 2e-5, 1e-4\} and batch size \{32, 64, 128\} for ENTROPY strategy. The best result over devset is achieved at 2e-5 learning rate and 32 batch size. 

Our BERT implementation is provided by Hugging Face and we choose "bert-base-uncased" version which contains 110M parameters and does not make a difference between lowercase and uppercase words.\footnote{\url{https://huggingface.co/bert-base-uncased}} We use the Adam optimizer with 0.001 weight decay \citep{loshchilov2017decoupled}. The size of the linear layer is 768 which is the same size of BERT final hidden state. Dropout with a probability of 0.3 is applied in the network. Training framework is based on Pytorch Lightning (version 1.5.8) which could greatly boosts training efficiency. All experiments use this network architecture.

\begin{table*}
\centering
\begin{tabular}{lll}
\hline
\textbf{Category} & \textbf{Definition} & \textbf{Example} \\
\hline
Emotion & \begin{tabular}[c]{@{}l@{}}  Plutchik's wheel of emotions \citep{plutchik1980general}.  \end{tabular} & \begin{tabular}[c]{@{}l@{}}  favored, scorned, frisky, \\ trustworthy, gripping, stilted  \end{tabular} \\
\hline
Need & \begin{tabular}[c]{@{}l@{}}  Maslow's hierarchy of needs \citep{maslow1943theory}.  \end{tabular} & \begin{tabular}[c]{@{}l@{}}  devout, deserving, protective, \\ hired, wealthy, rewarding  \end{tabular} \\
\hline
Perceiving & \begin{tabular}[c]{@{}l@{}}  individuals use information to form perceptions \\ of themselves and others based on  \\ social categories. \citep{alan2011perception}.  \end{tabular} & \begin{tabular}[c]{@{}l@{}} sensuous, ubiquitous, instinctive,  \\ detected, recognized, perceivable  \end{tabular} \\
\hline
Reasoning & \begin{tabular}[c]{@{}l@{}}  Deduction based on classical \\ logic \citep{shapiro2000classical} or mental \\ models \citep{johnson2001mental}.  \end{tabular} &  \begin{tabular}[c]{@{}l@{}}  suitable, predominate, critical, \\ substandard, relevant, causal  \end{tabular} \\
\hline
Planning & \begin{tabular}[c]{@{}l@{}}  Mental time travel \\ \citep{suddendorf2007evolution}. \end{tabular} & \begin{tabular}[c]{@{}l@{}}  committed, aimless, exploited, \\ unplanned, purposeful, executed  \end{tabular} \\
\hline
Personality & \begin{tabular}[c]{@{}l@{}}  Stable patterns of behavior, cognition, \\ and emotion \citep{corr2020cambridge}.  \end{tabular} &\begin{tabular}[c]{@{}l@{}}  whimsical, squeamish, shy, \\ punctual, entrepreneurial, intrepid  \end{tabular} \\
\hline
\end{tabular}
\caption{\label{tab:ob_word_mental_func2}
\emph{Mental} adjective examples which are classified as objective in subjectivity classification. The definition of each mental category is provided.  
}
\end{table*}

Each strategy is run three times with different random seeds and the averaged F1 scores over testset after three, four, five iterations are reported in Table~\ref{tab:f1}. ENTROPY outperforms the other three, achieving the highest \emph{Mental} F1 0.72 and \emph{Physical} F1 0.87 at iteration 4. The reason that CAL fails might be we don't find semantically similar neighbors as the size of $\mathcal{D}_{labeled}$ is too small. Table~\ref{tab:anno_num} shows annotation resource consumption. ENTROPY requires 60\textasciitilde70 labeled words per iteration, which means totally only 300 labeled words are needed to deliver an applicable classifier. CORESET and Random need more annotations than ENTROPY. CAL could not provide enough positive and negative samples after 120 words are annotated for some iterations. Precision and recall scores are presented in Appendix~\ref{sec:appendix_pr}.

\begin{table}[h]
\centering
\begin{tabular}{ccc}
\hline
\textbf{Strategy} & \textbf{Words/Iter} & \textbf{EnoughSamples}\\
\hline
ENTROPY & 60 \textasciitilde \ 70 & Yes \\
CORESET& 80 \textasciitilde \ 120 & Yes \\
CAL & 50 \textasciitilde \ 120 & No \\
Random & 80 \textasciitilde \ 100 & Yes \\
\hline
\end{tabular}
\caption{\label{tab:anno_num}
Range of annotation number per iteration. ENTROPY requires the lowest annotation resource. CAL could not provide enough positive and negative samples after 120 words are annotated for some iterations. 
}
\end{table}

\section{Comparison with SentiWordNet}
As the notion of \emph{Mental} and \emph{Physical} is to some extent similar to "subjective" and "objective" in the subjectivity classification task \citep{liu2010sentiment} of sentiment analysis, we'd like to investigate the difference between them. We choose to compare our result with SentiWordNet \citep{sebastiani2006sentiwordnet, baccianella2010sentiwordnet} which is the most used lexicon in social opinion mining studies \citep{cortis2021over}. SentiWordNet is a lexical resource which labels each synset from WordNet \citep{miller1995wordnet} as "positive", "negative" or "neutral". The used version of SentiWordNet is 3.0, which is based on WordNet 3.0. 

SentiWordNet 3.0 associates each synset with three numerical scores \emph{PosScore}, \emph{NegScore} and \emph{ObjScore} which show how positive, negative, and neutral the words contained in the synset are \citep{baccianella2010sentiwordnet}. All three scores range from 0 to 1 and their sum is 1. We focus on adjective synsets and classify each of them into two classes: \emph{SubSyn}, if the maximum of the three scores is \emph{PosScore} or \emph{NegScore}; otherwise, \emph{ObjSyn}. For an adjective that belongs to more than one synset, it owns different senses, perhaps having both subjective and objective meanings. Therefore, at lexical level, an adjective is labeled by this rule: \emph{Subjective}, if it only belongs to \emph{SubSyn} synsets;  \emph{Objective}, if it only belongs to \emph{ObjSyn} synsets; \emph{Dual}, if if belongs to both \emph{SubSyn} and \emph{ObjSyn} synsets. Table~\ref{tab:mental_senti_dist} shows the distribution of \emph{Subjective}, \emph{Objective} and \emph{Dual} adjectives in \emph{Mental} and \emph{Physical} classes. We find that 43\% of the \emph{Mental} adjectives are labeled as \emph{Objective}. This indicates the notions of \emph{Mental/Physical} are different from \emph{Subjective/Objective}. In fact, many \emph{Objective} adjectives bear mental functionalities. Some adjective examples are listed to illustrate this point in Table~\ref{tab:ob_word_mental_func2} in six categories: Emotion, Need, Perceiving, Reasoning, Planning and Personality.

\begin{table}
\centering
\resizebox{!}{0.75cm}{
\begin{tabular}{|c|c|c|c|c|}
\hline
\textbf{Class} & \textbf{Subjective} & \textbf{Objective} & \textbf{Dual} & \textbf{Total}\\
\hline
Mental & 28\% & 43\% & 29\% & 100\% \\
\hline
Physical & 9\% & 74\% & 17\% & 100\% \\
\hline
\end{tabular}
}
\caption{\label{tab:mental_senti_dist}
Distribution of \emph{Subjective}, \emph{Objective} and \emph{Dual} adjectives in \emph{Mental} and \emph{Physical} classes. Only 28\% of those words in \emph{Mental} fall into \emph{Subjective}, while 43\% belong to \emph{Objective}. This indicates many \emph{Objective} adjectives bear mental functionalities under MPC definition.
}
\end{table}

\section{Conclusion}
Aiming to explicitly reveal reasoning path in commonsense scenarios, our first step is to classify a word into \emph{Mental} or \emph{Physical}. We provide clear definitions of these two categories and a simple criterion for judging them. Active learning algorithm is implemented to fine-tune a BERT model, reducing the required annotation resources. The BERT model automatically infers which class an adjective belongs to. We release the inferred tags publicly to facilitate future research. We also compare our result with SentiWordNet, and find the notions of \emph{Mental/Physical} is different from \emph{Subjective/Objective} in sentiment analysis. Many \emph{Objective} adjectives bear mental functionalities under MPC definition.

Future research works focus on designing more fine-grained tags and training models to automatically infer them over words. Links are built between tags in a manual way or machine-learning style, to form an applicable reasoning graph. We hope to translate what humans know about the world and about themselves into graph that will improve the intelligence of machines. Large language models (LLMs) provide powerful techniques to extract data patterns in nature language, which makes it possible to perfectly associate words with all kinds of human-designed tags. However, at the level of reasoning, relying on LLMs is not necessarily feasible, and painstaking manual work may be essential.

\section*{Limitations}
\label{limit}
Although the model by ENTROPY achieves acceptable F1 scores, there's still a lot of room for improvement of classification precision and recall. For example, use more annotated words for fine-tuning, or try other deep learning algorithms. We leave this optimization in the future after we verify the whole research plan becomes feasible and the classification performance is a bottleneck for commonsense reasoning ability.

As a word has different meanings in different contexts, the best granularity for MPC is gloss level rather than lexical level. That's to say, use each piece of gloss text as BERT input instead of merging all glosses of a word into one piece of text. Then, the output shows if a gloss belongs to \emph{Mental} or \emph{Physical}. However, lexical level facilitate the development of reasoning graph as there's no need to consider context. We will change to gloss level by the time it's verified that context becomes a bottleneck and it should be integrated into reasoning graph.

\section*{Acknowledgements}
We appreciate valuable suggestions for this work from every reviewer.

\bibliography{custom}
\bibliographystyle{plain}

\appendix

\section{Word examples in MPC task}
\label{sec:appendix_infersub_example}
Table~\ref{tab:infersub_examples} shows seven adjectives with their classes in MPC task. From these examples, we could see clear difference in definition texts between \emph{Mental} and \emph{Physical} classes. Therefore, it's possible to fine-tune a high-accuracy BERT for MPC task.

\begin{table*}
\centering
\begin{tabular}{lll}
\hline
\textbf{Word} & \textbf{Class} & \textbf{Definition} \\
\hline
interested & \emph{Mental} & \begin{tabular}[c]{@{}l@{}} having or showing interest; especially curiosity or fascination or concern; \\ involved in or affected by or having a claim to or share in; \end{tabular} \\
\hline
angry & \emph{Mental} & \begin{tabular}[c]{@{}l@{}} feeling or showing anger; (of the elements) as if showing violent anger; \\severely inflamed and painful;  \end{tabular}\\
\hline
clever & \emph{Mental} & \begin{tabular}[c]{@{}l@{}} showing self-interest and shrewdness in dealing with others; \\mentally quick and resourceful; \end{tabular} \\
\hline
lazy & \emph{Mental} & \begin{tabular}[c]{@{}l@{}} moving slowly and gently; disinclined to work or exertion; \end{tabular}\\
\hline
molecular & \emph{Physical} & \begin{tabular}[c]{@{}l@{}} relating to or produced by or consisting of molecules; \\relating to simple or elementary organization;  \end{tabular} \\
\hline
blue & \emph{Physical} & \begin{tabular}[c]{@{}l@{}} of the color intermediate between green and violet; \\having a color similar to that of a clear unclouded sky;  \end{tabular}\\
\hline
automated & \emph{Physical} & \begin{tabular}[c]{@{}l@{}} operated by automation; \end{tabular} \\

\hline
\end{tabular}
\caption{\label{tab:infersub_examples}
Examples of words and their classes in MPC task. Definitions are provided by WordNet. From these examples, we could see clear difference in definition texts between two classes.
}
\end{table*}

\section{Precision and Recall}
\label{sec:appendix_pr}
Table~\ref{tab:precision_recall} shows averaged precision and recall after 3,4,5 iterations of each strategy. For \emph{Mental} class, ENTROPY achieves the highest precision around 0.8 and CAL has the highest recall above 0.9. For \emph{Physical} class, CAL achieves the highest precision above 0.9 and ENTROPY has the highest recall around 0.9.  

\begin{table*}
    \begin{subtable}{.5\linewidth}
      \centering
        \caption{\emph{Mental} Precision}
        \resizebox{!}{0.85cm}{
        \begin{tabular}{ccccc}
            \hline
            \textbf{Iteration} & \textbf{ENTROPY} & \textbf{CORESET} & \textbf{CAL} & \textbf{Random} \\
            \hline
            3 & 0.70 & 0.62 & 0.53 & 0.54 \\
             4 & \textbf{0.81} & 0.63 & 0.51 & 0.72 \\
             5 & \textbf{0.79} & 0.59 & 0.54 & 0.69 \\
	    \hline	
        \end{tabular}
        }
    \end{subtable}
    \begin{subtable}{.5\linewidth}
      \centering
        \caption{\emph{Mental} Recall}
        \resizebox{!}{0.85cm}{
        \begin{tabular}{ccccc}
            \hline
            \textbf{Iteration} & \textbf{ENTROPY} & \textbf{CORESET} & \textbf{CAL} & \textbf{Random} \\
            \hline
            3 & 0.54 & 0.79 & 0.74 & 0.73 \\
             4 & 0.65 & 0.81 & \textbf{0.87} & 0.57 \\
             5 & 0.64 & 0.82 & \textbf{0.94} & 0.61 \\
	    \hline		
        \end{tabular}
        }
    \end{subtable} 
    \begin{subtable}{.5\linewidth}
      \centering
        \caption{\emph{Physical} Precision}
        \resizebox{!}{0.85cm}{
        \begin{tabular}{ccccc}
            \hline
            \textbf{Iteration} & \textbf{ENTROPY} & \textbf{CORESET} & \textbf{CAL} & \textbf{Random} \\
            \hline
            3 & 0.77 & 0.86 & 0.81 & 0.79 \\
             4 & 0.82 & 0.87 & \textbf{0.88} & 0.79 \\
             5 & 0.82 & 0.88 & \textbf{0.94} & 0.79 \\
	    \hline		
        \end{tabular}
        }
    \end{subtable} 
    \begin{subtable}{.5\linewidth}
      \centering
        \caption{\emph{Physical} Recall}
        \resizebox{!}{0.85cm}{
        \begin{tabular}{ccccc}
            \hline
            \textbf{Iteration} & \textbf{ENTROPY} & \textbf{CORESET} & \textbf{CAL} & \textbf{Random} \\
            \hline
            3 & 0.87 & 0.73 & 0.64 & 0.60 \\
             4 & \textbf{0.91} & 0.73 & 0.53 & 0.88 \\
             5 & \textbf{0.90} & 0.68 & 0.55 & 0.83 \\
	    \hline	
        \end{tabular}
        }
    \end{subtable} 
    \caption{Averaged precision and recall of \emph{Mental} and \emph{Physical} after 3,4,5 iterations.} 
    \label{tab:precision_recall}
\end{table*}

\end{document}